\documentclass[conference]{IEEEtran}
\IEEEoverridecommandlockouts

\usepackage{multirow}
\usepackage{lipsum}
\usepackage{amsmath}
\usepackage{amssymb}
\usepackage[ruled,vlined]{algorithm2e}
\usepackage{graphicx}
\graphicspath{{./Figures/}}
\usepackage{amsmath}
\usepackage{graphicx}  
\usepackage{subfig}

\usepackage{multirow}
\usepackage{booktabs}
\usepackage{xcolor}
\usepackage{float}
\usepackage{cite}
\usepackage{stfloats}  
\usepackage[font=footnotesize]{caption}
\usepackage{mathtools} 

\usepackage{url}

\def\BibTeX{{\rm B\kern-.05em{\sc i\kern-.025em b}\kern-.08em
    T\kern-.1667em\lower.7ex\hbox{E}\kern-.125emX}}
    
\begin{document}

\title{Online Joint Calibration of Steering Offset and Planar LiDAR Extrinsics for Wheeled Mobile Robots
}

\author{Subodh Mishra, Arindam Dhar, Suprotim Majumdar, Naveen Arulselvan
\thanks{All authors are with ATI Motors, Bangalore, India}
}


\maketitle

\begin{abstract}
Accurate steering sensing and LiDAR-to-vehicle extrinsics are crucial for reliable path tracking in warehouse mobile robots (WMRs); miscalibration often leads to snaking, weaving, and elevated cross-track error (CTE). In practice, steering ``zero'' is commonly set manually (e.g., eyeballing straightness via a PS4 joystick), while LiDAR extrinsics are assumed from CAD and may drift after maintenance. Such static, manual procedures frequently cause  miscalibration in safety-critical environments. This paper presents an Extended Kalman Filter (EKF)--based method for online estimation of steering offset and planar LiDAR extrinsics within a bicycle-kinematics model, providing a principled alternative to manual calibration. Experiments on real datasets show that correcting steering offset reduces CTE substantially, validating the effectiveness of the proposed approach.
\end{abstract}

\vspace{-5mm}
\section{Introduction}
\label{sec: introduction}
Accurate and safe control response is fundamental to reliable operation of autonomous warehouse mobile robots (WMRs) that rely on fusion of wheel-encoder kinematics, steering angle sensing, and LiDAR-based odometry to estimate their motion in real time. Sensor fusion depends on: (i) the mapping between measured and true steering angle, and (ii) the planar extrinsic transform between the LiDAR and the kinematic control point. Imperfections in either can introduce systematic drift, trajectory/map distortion, and elevated cross-track error (CTE) during closed loop control which are unacceptable in safety-critical warehouse environments. Our current industrial practice remains largely manual. Steering ``zero'' is typically aligned visually by using a PS4 joystick to center the wheels and/or by manually driving the robot forward and backward to approximate straightness. The full planar LiDAR extrinsics are typically fixed to CAD or one-time measurements, even though the vehicle’s kinematic center lies on the body while the LiDAR is mounted on the chassis; mounting flex, vibration, or re-assembly can cause relative motion between these structures and thus alter the true extrinsics over time.

To overcome these limitations, we propose an Extended Kalman Filter (EKF) based method for online estimation of steering angle offset and full planar LiDAR extrinsic calibration within a bicycle-model vehicle. We augment the state with the LiDAR extrinsics \((x_{\text{off}}, y_{\text{off}}, \theta_{\text{off}})\) and an additive steering bias $\delta_{off}$, and continuously estimate these parameters by fusing bicycle model kinematics with LiDAR odometry. We demonstrate that the proposed approach correctly detects steering miscalibration, converges to reliable values of extrinsics and improves Cross Track Error during closed loop control.
\begin{figure}[htbp]
    \vspace{-5mm}
    \centering

    \subfloat[Tugger WMR\label{fig:atitug}]{
        \includegraphics[width=0.48\linewidth]{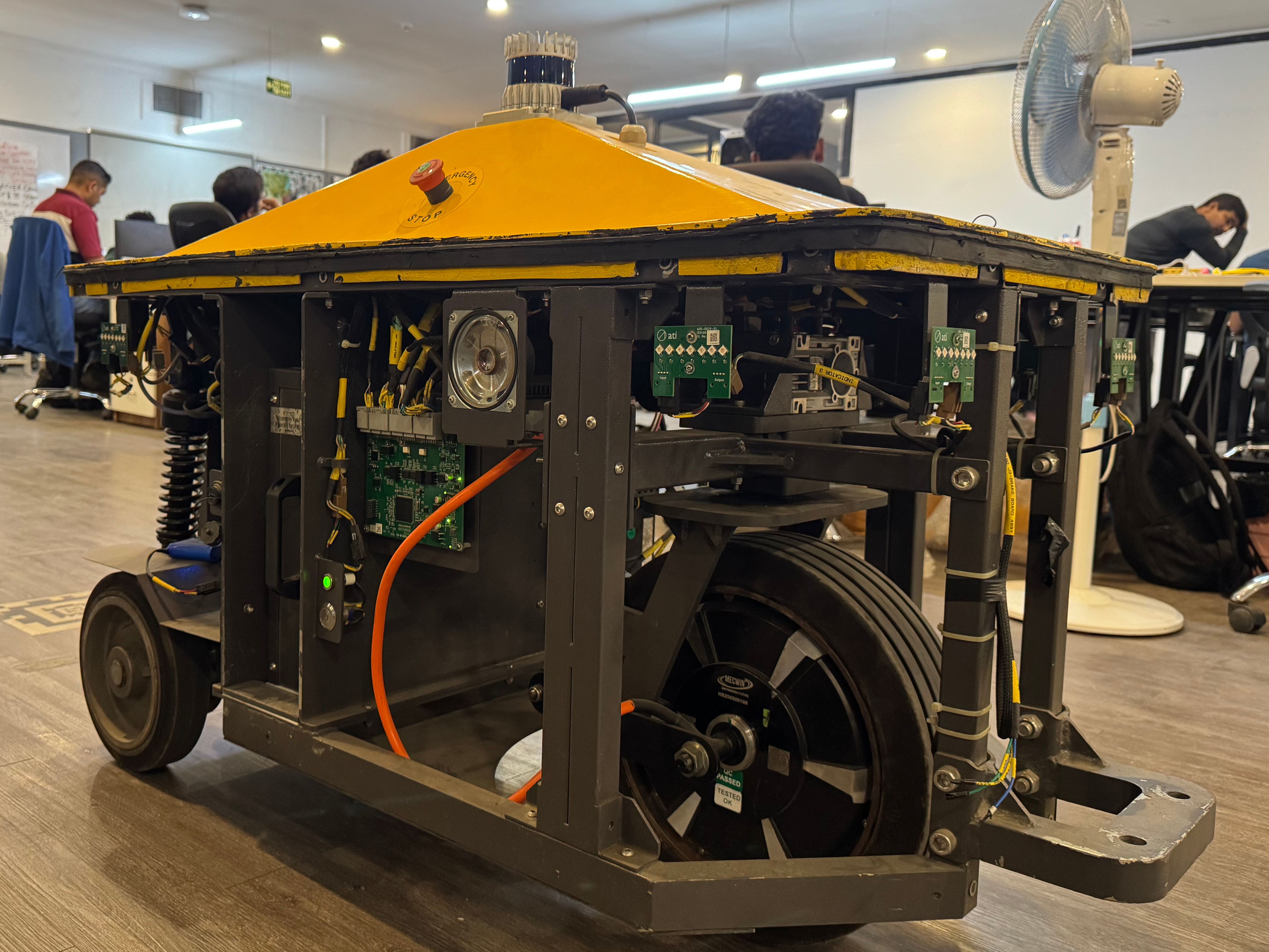}
    }
    \hfill
    \subfloat[Nano WMR\label{fig:atinano}]{
        \includegraphics[width=0.44\linewidth]{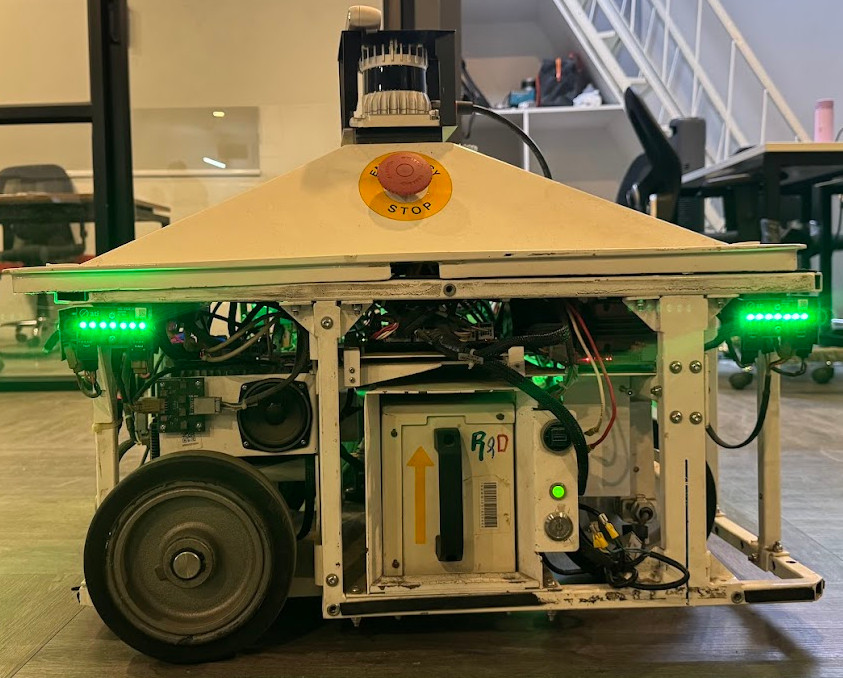}
    }

    \caption{\textbf{Experimental WMR Platforms}. 
    (a) An autonomous tugger that hauls trailers/payloads in industrial sites.
    (b) Experimental platform used for validation and benchmarking of autonomy algorithms.
    Both platforms use an Ouster 32-channel LiDAR running at 10 Hz. The 3D point cloud is converted to 2D range scan after filtering out moving objects, the floor and the ceiling and then by setting the z of all the remaining points to zero.
    Encoder-derived longitudinal velocity and steering angle arrive at 100 Hz.}
    \label{fig:wmr_platforms}
    \vspace{-5mm}
\end{figure}


\section{Related Work}
\label{sec: relatedwork}
\cite{borenstein1996} introduced one of the earliest systematic procedures for detecting and correcting kinematic biases in an offline fashion in differential drive mobile robots via the bi-directional “UMBmark’’ test. By analyzing return-position errors from clockwise and counter-clockwise square-path runs, they isolated turning and curvature biases and compensated them through corrected differential-drive kinematics. \cite{roy1999} proposed one of the first online self-calibration approaches, estimating systematic odometric drift (and not robot parameters) in differential drive robots by aligning consecutive laser scans and updating a linear drift model under a maximum-likelihood framework. \cite{siegwart2003a} present an approach to estimate both systematic and non-systematic odometry errors by combining an Augmented Kalman Filter (AKF) with an Observable Filter (OF). Systematic terms such as wheel-radius mismatches, wheelbase uncertainty, and rotation or steering scale factors are estimated online, while non-systematic disturbances from slip or surface irregularities are inferred through observables derived from consecutive pose estimates. Their framework supports differential and synchronous drives but is evaluated only in simulation. 

\cite{eliazar2004} propose to learn a probabilistic motion model by coupling a particle-filter SLAM algorithm with an EM procedure. Although effective, this method is computationally intensive and cannot operate online. \cite{martinelli2006} introduce an EKF-based method to estimate the extrinsic pose of an onboard omnidirectional vision sensor on a differential-drive robot using only wheel odometry and bearing measurements to a fixed landmark. \cite{underwood2007} present a batch optimization technique for estimating the full 6-DoF extrinsic pose of a range sensor by driving around a known calibration structure (vertical pole and ground plane) and minimizing geometric residuals. Their method relies on a known environment and global pose measurements. 

\cite{censi2008} provide a closed-form batch/offline maximum-likelihood technique for simultaneously calibrating differential-drive odometry parameters (wheel radii and wheel baseline) and planar LiDAR extrinsics using only wheel encoder velocities and scan-matching–derived relative motions.  \cite{kummerle2011} extend graph-based SLAM to jointly estimate the robot trajectory, the map, LiDAR extrinsics, and differential-drive odometry parameters by treating calibration variables as nodes in a hyper-graph. Their sliding-window optimization supports online adaptation to parameter drift (e.g., load-dependent wheel radii) paving the path for long term calibration, localization and mapping. While the methods so far focussed primarily on differential drive systems, \cite{kallasi2017} present an intrinsic–extrinsic calibration framework for industrial tricycle (bicycle-equivalent) AGVs, jointly estimating steering offset, driving scale, and planar LiDAR pose using encoder data and landmark-based LiDAR egomotion. This work comes closest to our problem setting. However, it relies on batch, quasi–closed-form estimation from repeated constant-curvature trajectories and requires an externally referenced navigation scanner.

\textbf{Contributions.}
In contrast to prior work targeting differential-drive platforms 
\cite{borenstein1996,roy1999,siegwart2003a,eliazar2004,censi2008,kummerle2011} 
or relying on controlled circular paths and external navigation references 
\cite{kallasi2017}, our method performs online joint estimation of steering offset and planar LiDAR extrinsics on a bicycle-modelled WMR during natural motion, without any external aids. 
Although no pre-scripted trajectory is required, sufficient motion excitation remains necessary due to inherent unobservabilities. 
A further contribution is the use of motion-based calibration constraint within the state-estimation formulation; while similar ideas have appeared in manipulator~\cite{Horaud1995}, LiDAR–camera~\cite{taylor2016}, and LiDAR–IMU~\cite{mishra2021} calibration, we are not aware of prior work applying them to jointly calibrate steering and planar LiDAR extrinsics on a WMR. 
Finally, we quantify how the recovered steering offset improves cross-track error (CTE) - a critical evaluation often omitted in calibration literature, thus validating the complete calibration–to–control pipeline.
\section{Bicycle Kinematic Model}
\label{sec: bicyclekinematicmodel}
We consider our WMR modeled using the standard planar bicycle kinematic model. Let the robot state at time $k$ include the vehicle pose 
$(x_k, y_k, \theta_k)$ expressed in a global frame, where $x_k$ and $y_k$ denote the position of the rear axle center and $\theta_k$ denotes the vehicle heading, at time $t_k$. The robot is actuated through a longitudinal velocity $v_k$ obtained from wheel-encoder measurements and a measured steering input $\delta^{\mathrm{meas}}_k$ obtained from the steering encoder.

\subsection{Steering Bias and Motion Model}
\label{sec:steering_and_motion}

The measured steering angle is typically corrupted by a slowly varying bias arising from mechanical misalignment, wear, or encoder inconsistencies. We model the true steering angle as
\begin{equation}
    \delta^{\mathrm{true}}_k = \delta^{\mathrm{meas}}_k - \delta_{off,k},
\end{equation}
where $\delta_{off,k}$ is an unknown constant (or slow varying) offset to be estimated.

Under the bicycle-kinematics assumption and small-slip conditions, the vehicle pose evolves as (\cite{Luca1995})
\begin{equation}
\label{eq:bicycle_ct}
\begin{aligned}
    \dot{x}(t) &= v(t)\cos\theta(t),\\
    \dot{y}(t) &= v(t)\sin\theta(t),\\
    \dot{\theta}(t) &= \frac{v(t)}{L}\tan\!\bigl(\delta^{\mathrm{true}}(t)\bigr),
\end{aligned}
\end{equation}
where $L$ is the wheelbase. Applying a zero-order hold on the inputs over sampling time of $\Delta t_k$ yields the discrete-time update
\begin{equation}
\label{eq:bicycle_discrete}
\begin{aligned}
    x_{k+1} &= x_k + v_k\cos\theta_k\,\Delta t_k, \\
    y_{k+1} &= y_k + v_k\sin\theta_k\,\Delta t_k, \\
    \theta_{k+1} &= \theta_k 
        + \frac{v_k}{L}\tan\!\bigl(\delta^{\mathrm{meas}}_k - \delta_{off,k}\bigr)\Delta t_k.
\end{aligned}
\end{equation}
This formulation makes the steering offset directly influence the predicted curvature. While many implementations treat bicycle-type vehicles as unicycles by using the yaw rate~$\omega$ as the control input, thus bypassing the $v\tan(\delta)/L$ relation and yielding a control-affine model, this simplification obscures the coupling between steering miscalibration and trajectory curvature, making the bias unobservable. Retaining the full bicycle model preserves this coupling and maintains observability of the steering offset.

\subsection{State Augmentation}
\label{sec: stateaugmentation}
To enable online calibration, the state vector is augmented to include the planar LiDAR-to-vehicle extrinsic parameters $
    \mathbf{e} = 
    \begin{bmatrix}
        x_{\mathrm{off}} & y_{\mathrm{off}} & \theta_{\mathrm{off}}
    \end{bmatrix}^{\top}$
and the steering bias $\delta_{off}$. These quantities evolve according to a random-walk model,
\begin{equation}
\label{eq:calib_random_walk}
\begin{aligned}
    \mathbf{e}_{k+1} &= \mathbf{e}_{k} + \mathbf{w}^{\mathrm{ex}}_k, \\
    \delta_{off,\,k+1} &= \delta_{off,\,k} + w^{\delta}_k.
\end{aligned}
\end{equation}
where $\mathbf{w}^{\mathrm{ex}}_k$ and $w^{\delta}_k$ denote small process noise terms that allow slow adaptation to mechanical drift or reassembly-induced changes. The augmented state vector becomes $\mathbf{s}_k =
    \begin{bmatrix}
        x_k & y_k & \theta_k &
        \mathbf{e}_{k}^{\top} &
        \delta_{off}
    \end{bmatrix}^{\top}$.
\subsection{Process Model Jacobian}
\label{sec: processmodeljacobian}
\begin{equation}
\mathbf{F}_k = 
\left.\dfrac{\partial f}{\partial \mathbf{s}}\right|_{(\hat{\mathbf{s}}_k,\mathbf{u}_k)},
\qquad
\mathbf{G}_k = 
\left.\dfrac{\partial f}{\partial \mathbf{u}}\right|_{(\hat{\mathbf{s}}_k,\mathbf{u}_k)}.
\label{eqn: processjacobianeqn}
\end{equation}
For EKF propagation, the motion model $f(\mathbf{s_k}, \mathbf{u_k})$ is linearized about the current estimate 
$\hat{\mathbf{s}}_k$, yielding the state and input Jacobians $\mathbf{F}_k$ and $\mathbf{G}_k$ respectively (Equation \ref{eqn: processjacobianeqn}). The derivation skipped in the interest of space.
\section{Measurement Model}
\label{sec: measurementmodel}
The measurement model relates measurements to state variables. LiDAR odometry provides the measurement that is used to update the state variables. The motion based calibration constraint \cite{taylor2016}, \cite{mishra2021} is utilized to formulate the measurement model.
\vspace{-3.5mm}
\subsection{LiDAR Odometry}
\label{sec: lidarodometry}
2D LiDAR odometry is performed by estimating the sensor's pose through
scan–to–map alignment using a discrete grid search around the sensor's pose at the last timestamp. Starting from an initial pose prediction seeded by a known location in the map, the algorithm conducts a discrete grid search
over a bounded neighborhood in $x$, $y$, and $\theta$. A set of candidate transformations based on the expected max speed and direction is applied to
the filtered and randomly sampled incoming scan, and each candidate is evaluated by computing an
alignment score that measures how well the transformed scan points agree
with the occupied regions of the map. The best-scoring pose  is selected as the pose that maximizes the alignment score,
yielding a robust and repeatable scan-matching solution suitable for
real-time localization and mapping.
\vspace{-1.5mm}
\subsection{Motion Based Calibration Constraint}
\label{sec: mocal}
\begin{figure}[htbp]
    \centering
    \includegraphics[width=0.4\linewidth]{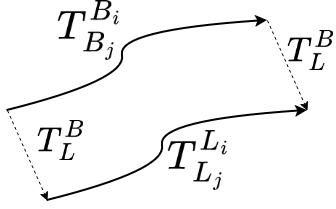}
    \caption{Motion-Based Calibration Constraint.}
    \label{fig: mocal}
    \vspace{-3mm}
\end{figure}
The motion-based calibration constraint (Figure \ref{fig: mocal}, Equation \ref{eq: mocal}, \cite{taylor2016}, \cite{mishra2021}) establishes a relationship between the motions of two coordinate frames rigidly attached to the body, using the extrinsic calibration ($T^{B}_{L}$) that connects them.
\begin{equation}
    T^{B_i}_{B_j} T^{B}_{L} = T^{B}_{L} T^{L_i}_{L_j}
    \label{eq: mocal}
\end{equation}
Note that $T^{A}_{B}\in SE(2)$ is a transformation matrix that stores the rotation and translation between frames A and B. In the present context $T^{B_i}_{B_j}$ is the motion estimated by vehicle kinematic model, $T^{L_i}_{L_j}$ is the LiDAR motion that can be estimated using LiDAR odometry. 

\subsection{Measurement Model}
\label{sec:measurement_model}
We use the relative LiDAR odometry estimate $T^{L_i}_{L_j}$ as the measurement.  
Starting from the motion constraint in~\eqref{eq: mocal}, setting $i=0$ and $j=k$
gives the EKF measurement model
\begin{equation}
    T^{L_0}_{L_k}
    =
    (T^{B}_{L})^{-1} (T^{W}_{B_0})^{-1}
    T^{W}_{B_k} \, T^{B}_{L},
    \label{eqn:measurementmodelekf}
\end{equation}
where $T^{B}_{L}=f(x_{off},y_{off},\theta_{off})$ and
$T^{W}_{B_k}=f(x_k,y_k,\theta_k)$, and $T^{L_0}_{L_k}=f(x_{L_k},y_{L_k},\theta_{L_k})$.
Equation~\eqref{eqn:measurementmodelekf} is therefore a function of the state $\mathbf{s}_k$.
and can be written compactly as $z_k = h(\mathbf{s}_k)$, where $z_k = T^{L_0}_{L_k}$.
Here $T^{W}_{B_0}$ is the fixed reference pose.

Expanding~\eqref{eqn:measurementmodelekf} yields the planar relative transform
{\small \begin{align*}
x_{L_k} &= (1-\cos\Delta\theta_k)x_{off}
          + \sin\Delta\theta_k\,y_{off} \\
        &\quad + \cos(\theta_{off}-\theta_0)\Delta x_k
          - \sin(\theta_{off}-\theta_0)\Delta y_k , \\[4pt]
y_{L_k} &= -\sin\Delta\theta_k\,x_{off}
          + (1-\cos\Delta\theta_k)y_{off} \\
        &\quad + \sin(\theta_{off}-\theta_0)\Delta x_k
          + \cos(\theta_{off}-\theta_0)\Delta y_k , \\[4pt]
\theta_{L_k} &= \Delta\theta_k .
\end{align*}

with $\Delta x_k=x_k-x_0$, $\Delta y_k=y_k-y_0$ and
$\Delta\theta_k=\theta_k-\theta_0$.

The measurement Jacobian is derived as \begin{align} \mathbf{H_k} &= \begin{bmatrix} \frac{\partial x_{L_k}}{\partial x_k} & \frac{\partial x_{L_k}}{\partial y_k} & \frac{\partial x_{L_k}}{\partial \theta_k} & \frac{\partial x_{L_k}}{\partial x_{off}}& \frac{\partial x_{L_k}}{\partial y_{off}}& \frac{\partial x_{L_k}}{\partial \theta_{off}}& \frac{\partial x_{L_k}}{\partial \delta_{off}} \\ \frac{\partial y_{L_k}}{\partial x_k} & \frac{\partial y_{L_k}}{\partial y_k} & \frac{\partial y_{L_k}}{\partial \theta_k} & \frac{\partial y_{L_k}}{\partial x_{off}}& \frac{\partial y_{L_k}}{\partial y_{off}}& \frac{\partial y_{L_k}}{\partial \theta_{off}}& \frac{\partial y_{L_k}}{\partial \delta_{off}}\\ \frac{\partial \theta_{L_k}}{\partial x_k} & \frac{\partial \theta_{L_k}}{\partial y_k} & \frac{\partial \theta_{L_k}}{\partial \theta_k} & \frac{\partial \theta_{L_k}}{\partial x_{off}}& \frac{\partial \theta_{L_k}}{\partial y_{off}}& \frac{\partial \theta_{L_k}}{\partial \theta_{off}}& \frac{\partial \theta_{L_k}}{\partial \delta_{off}} \end{bmatrix} \end{align}}
The model~\eqref{eqn:measurementmodelekf} and Jacobian $\mathbf{H}_k$ are used in
the EKF update step to minimize the innovation between the observed LiDAR
relative pose and its prediction.
\section{EKF Formulation}
\label{sec: ekfformulation}
\begin{figure}[H]
\vspace{-6.5mm}
    \centering
    \includegraphics[width=0.7\linewidth]{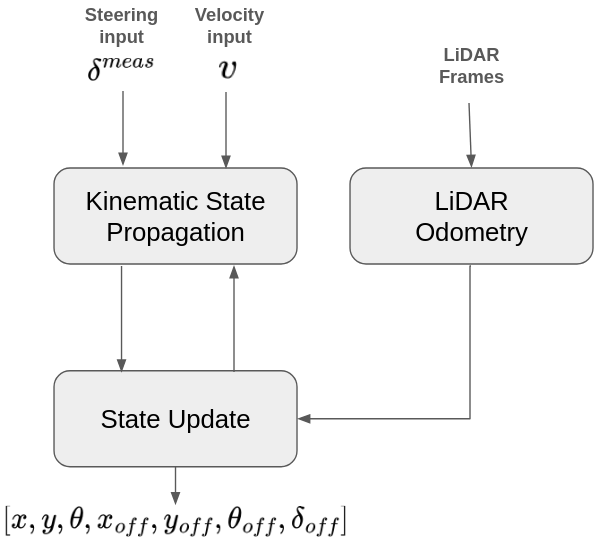}
    \caption{Full EKF pipeline.}
    \label{fig: fullpipelineekf}
\vspace{-4mm}
\end{figure}
Given the process model (Section \ref{sec: bicyclekinematicmodel}) and the measurement model (Section \ref{sec: measurementmodel}) derived from the motion based calibration constraint (Section \ref{sec: mocal}), online calibration is posed as an Extended
Kalman Filter (EKF) over the augmented state containing vehicle pose, planar LiDAR extrinsics, and steering bias. The EKF maintains a Gaussian belief
$(\hat{\mathbf{s}}_k,\mathbf{P}_k)$.
\subsection{Nonlinear Process Model}
\label{sec: nonlinearprocessmodel}
$\mathbf{u}_k=[v_k,\delta^{\mathrm{meas}}_k]^\top$ denotes the control input.
The dynamics are
\begin{equation}
    \mathbf{s}_{k+1}=f(\mathbf{s}_k,\mathbf{u}_k)+\mathbf{w}_k,
\end{equation}
with process noise $\mathbf{w}_k\sim\mathcal{N}(\mathbf{0},\mathbf{Q}_k)$
capturing velocity and steering uncertainty as well as slow drift of the
calibration parameters. Small random-walk noise is injected into
$(x_{\mathrm{off}},y_{\mathrm{off}},\theta_{\mathrm{off}},\delta_{\mathrm{off}})$
to allow gradual adaptation.
\subsection{State Prediction}
\label{sec: stateprediction}
Using the current estimate
$\hat{\mathbf{s}}_{k|k}$ , the input $\mathbf{u}_k$, and the process Jacobians
$\mathbf{F}_k$ and
$\mathbf{G}_k$ as described earlier in Section \ref{sec: processmodeljacobian}.  
The predicted state and covariance are
\begin{equation}
\label{eq:ekf_prediction}
\begin{aligned}
    \hat{\mathbf{s}}_{k+1|k} &= f(\hat{\mathbf{s}}_{k|k},\,\mathbf{u}_k), \\
    \mathbf{P}_{k+1|k} &= \mathbf{F}_k\,\mathbf{P}_{k|k}\,\mathbf{F}_k^\top
    + \mathbf{G}_k\,\mathbf{Q}_k\,\mathbf{G}_k^\top
\end{aligned}
\end{equation}
$\mathbf{Q}_k$ is tunable, block-diagonal process covariance matrix.
\subsection{Measurement Update}
\label{sec: measurementupdate}
Whenever a LiDAR-odometry pose is available, the EKF performs a correction using
the measurement model in~\eqref{eqn:measurementmodelekf} -- $\mathbf{z}_k = h(\mathbf{s}_k)$ --
which predicts the relative LiDAR pose implied by the current state estimate.
Using the most recent $(\hat{\mathbf{s}}_k,\mathbf{P}_k)$, the measurement Jacobian $\mathbf{H}_{k}$ and the tunable measurement covariance matrix $\mathbf{R}_{k}$, the innovation $\mathbf{r}_k$, its covariance $\mathbf{S}_{k}$, and the Kalman gain $\mathbf{K}_k$ are computed
\begin{equation}
\label{eq:ekf_update_stats}
\begin{aligned}
    \mathbf{r}_k &= \mathbf{z}_k - h(\hat{\mathbf{s}}_{k+1|k}), \\
    \mathbf{S}_k &= \mathbf{H}_k\,\mathbf{P}_{k+1|k}\,\mathbf{H}_k^\top + \mathbf{R}_k, \\
    \mathbf{K}_k &= \mathbf{P}_{k+1|k}\,\mathbf{H}_k^\top\,\mathbf{S}_k^{-1}.
\end{aligned}
\end{equation}
leading to the corrected estimates
\begin{equation}
\label{eq:ekf_correction}
\begin{aligned}
    \hat{\mathbf{s}}_{k+1|k+1} &= \hat{\mathbf{s}}_{k+1|k} + \mathbf{K}_k\,\mathbf{r}_k, \\
    \mathbf{P}_{k+1|k+1} &= (\mathbf{I}-\mathbf{K}_k\mathbf{H}_k)\,\mathbf{P}_{k+1|k}.
\end{aligned}
\end{equation}
The measurement noise $\mathbf{R}_k$ is obtained statistics of
the LiDAR odometry.
\vspace{-5mm}
\section{Experimental Evaluation}
\label{sec: experimentalevaluation}
We evaluate the proposed online calibration method on two WMRs (Figure~\ref{fig:wmr_platforms}) over several days. During this period, the platforms were actively used by other teams for daily testing and frequent re-assembly, naturally introducing unmodelled mechanical variations across datasets. In the absence of external ground-truth systems (e.g., GPS or total stations), we structure our empirical study around five validation tasks: 
(a)~assessing convergence of the steering-bias and planar LiDAR extrinsic estimates under different load and initial conditions; 
(b)~evaluating cross-platform generalization by deploying the algorithm on a second robot; 
(c)~quantifying the estimator’s sensitivity to deliberately induced steering miscalibration; 
(d)~examining the effect of re-applying and compensating the estimated steering offset on cross-track error (CTE); and 
(e)~studying the influence of steering-offset correction on the LiDAR odometry trajectory. 
The robot used for tasks (a), (c), (d), and (e) is shown in Figure~\ref{fig:atitug}, while task (b) uses the platform in Figure~\ref{fig:atinano}. In all experiments, unless specified otherwise, the LiDAR extrinsics are initialized to reference values obtained from tape measurements or CAD, and the steering offset is initialized to zero. Empirically, we observe that the onboard LiDAR odometry remains accurate to within approximately 10\,cm over the evaluated operating regions.

\subsection{Estimator Convergence}
\label{sec: estimatorconvergence}
\begin{figure}[htbp]
    \centering
    \subfloat[Unladen test\label{fig:calibxythetadelta_unladen}]{
        \includegraphics[width=\linewidth]{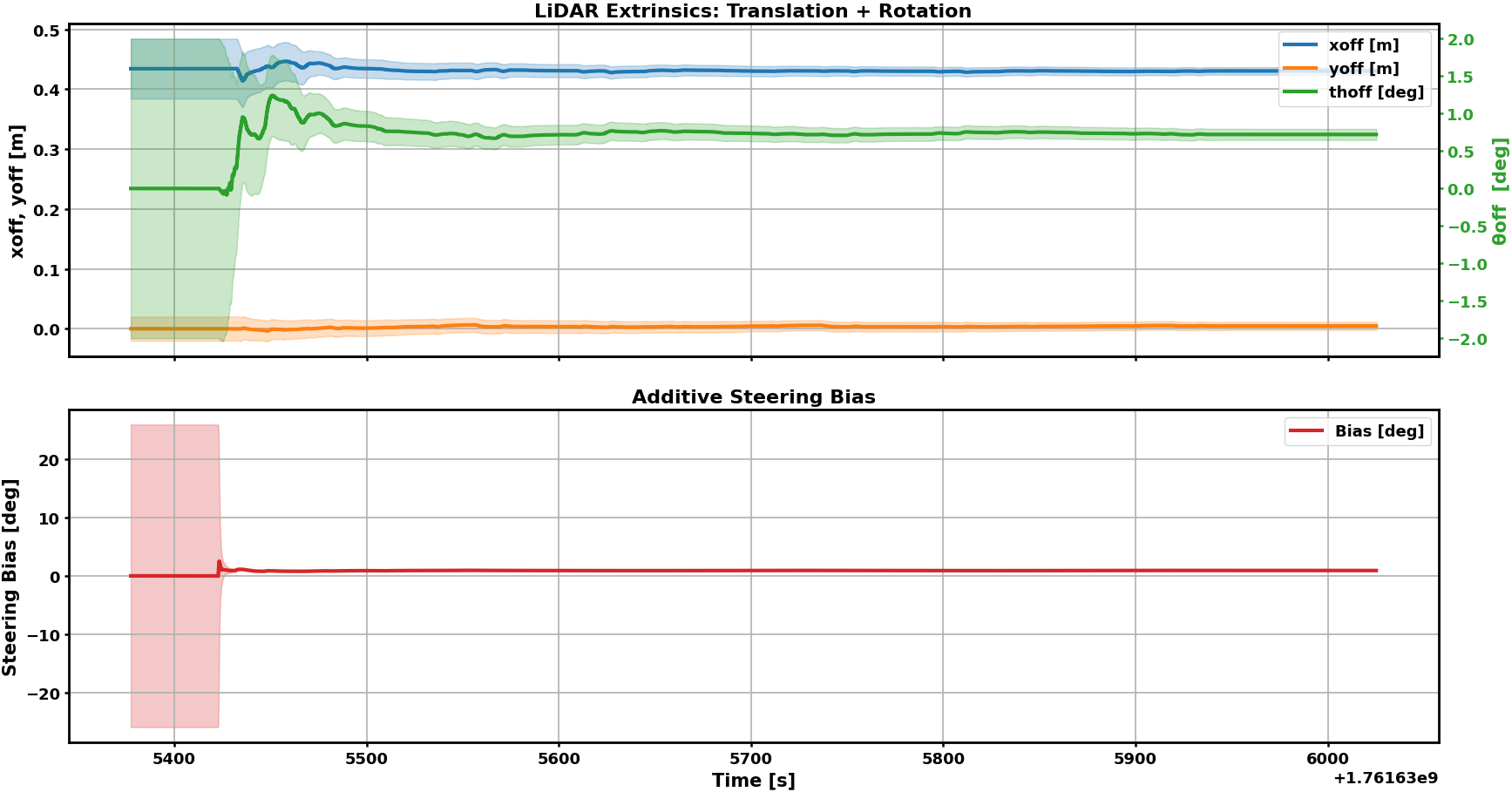}
    }
    \\[0.35cm]  
    \subfloat[Laden test with 500\,kg payload\label{fig:calibxythetadelta_laden}]{
        \includegraphics[width=\linewidth]{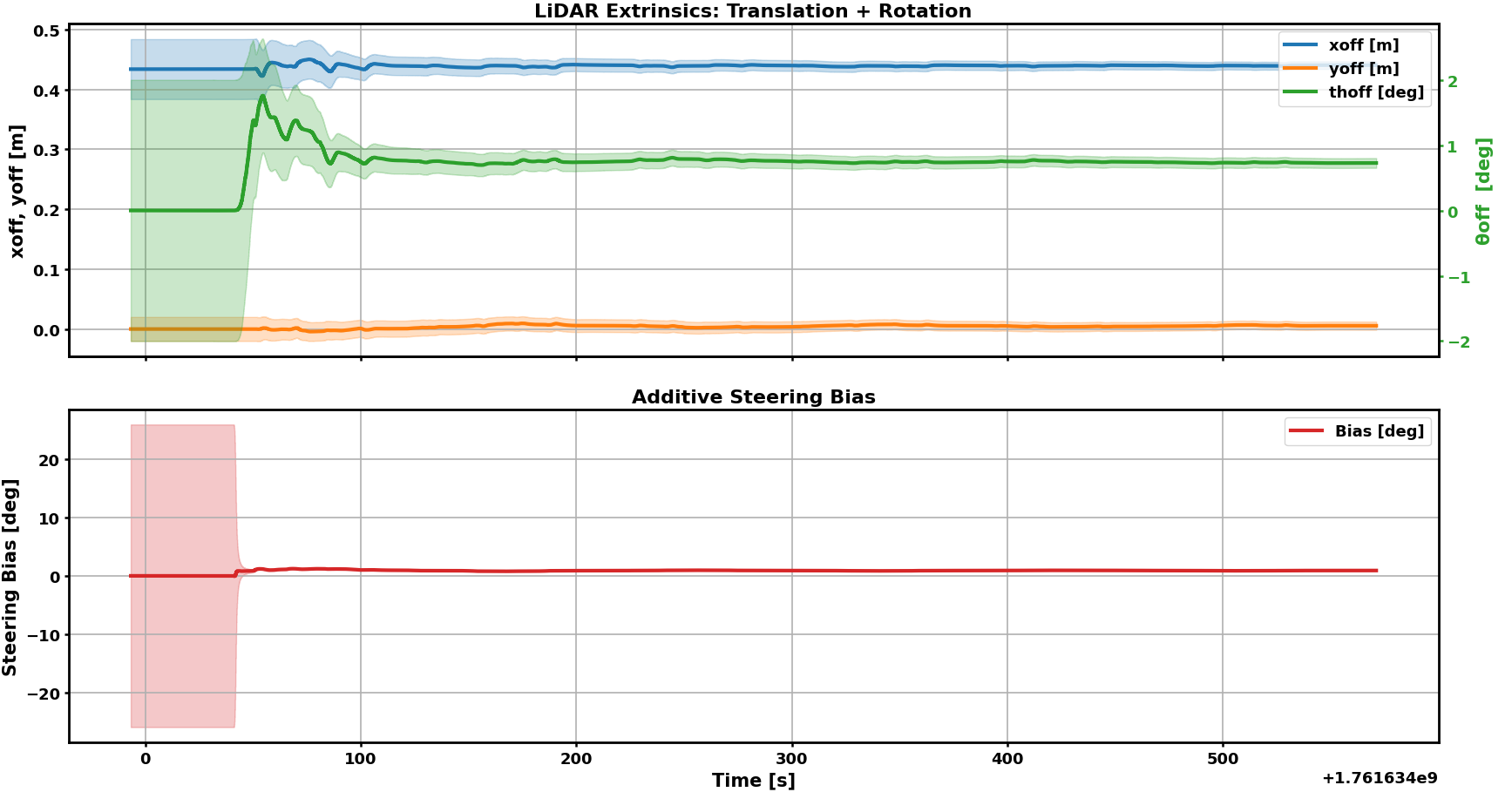}
    }
    \caption{\textbf{Estimator convergence.} In both load conditions, the estimator was initialized at CAD-provided values and converges to the steady-state values in Table~\ref{tab:ladenunladentest}. Note that the $\theta$ axis is on the right.}
    \label{fig:calibxythetadelta_combined}
    \vspace{-4mm}
\end{figure}

\begin{figure}[htbp]
    \centering

    \subfloat[Unladen test\label{fig:calibxythetadelta_unladen_zeroinit}]{
        \includegraphics[width=\linewidth]{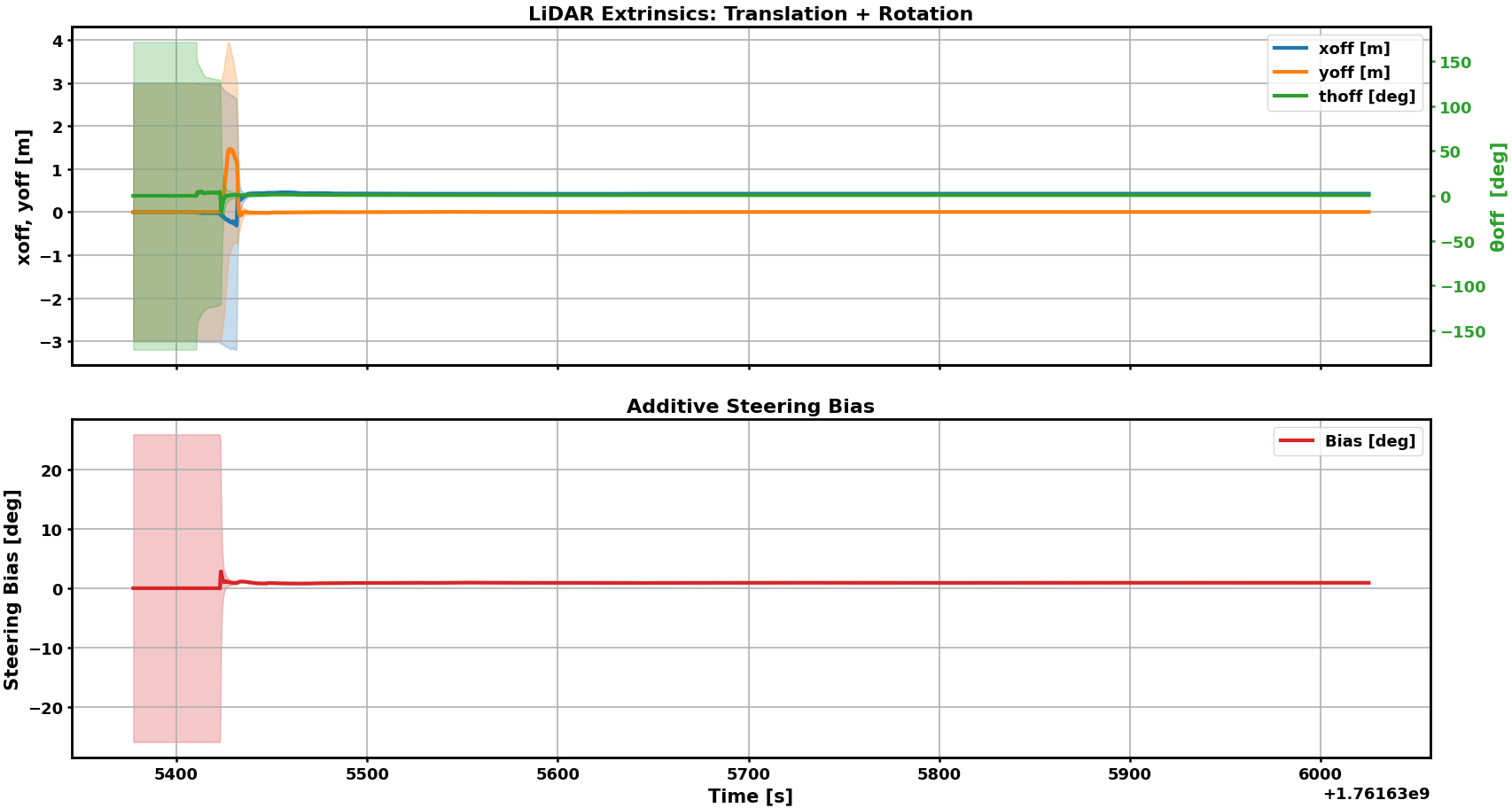}
    }
    \\[0.35cm]  

    \subfloat[Laden test with 500\,kg payload\label{fig:calibxythetadelta_laden_zeroinit}]{
        \includegraphics[width=\linewidth]{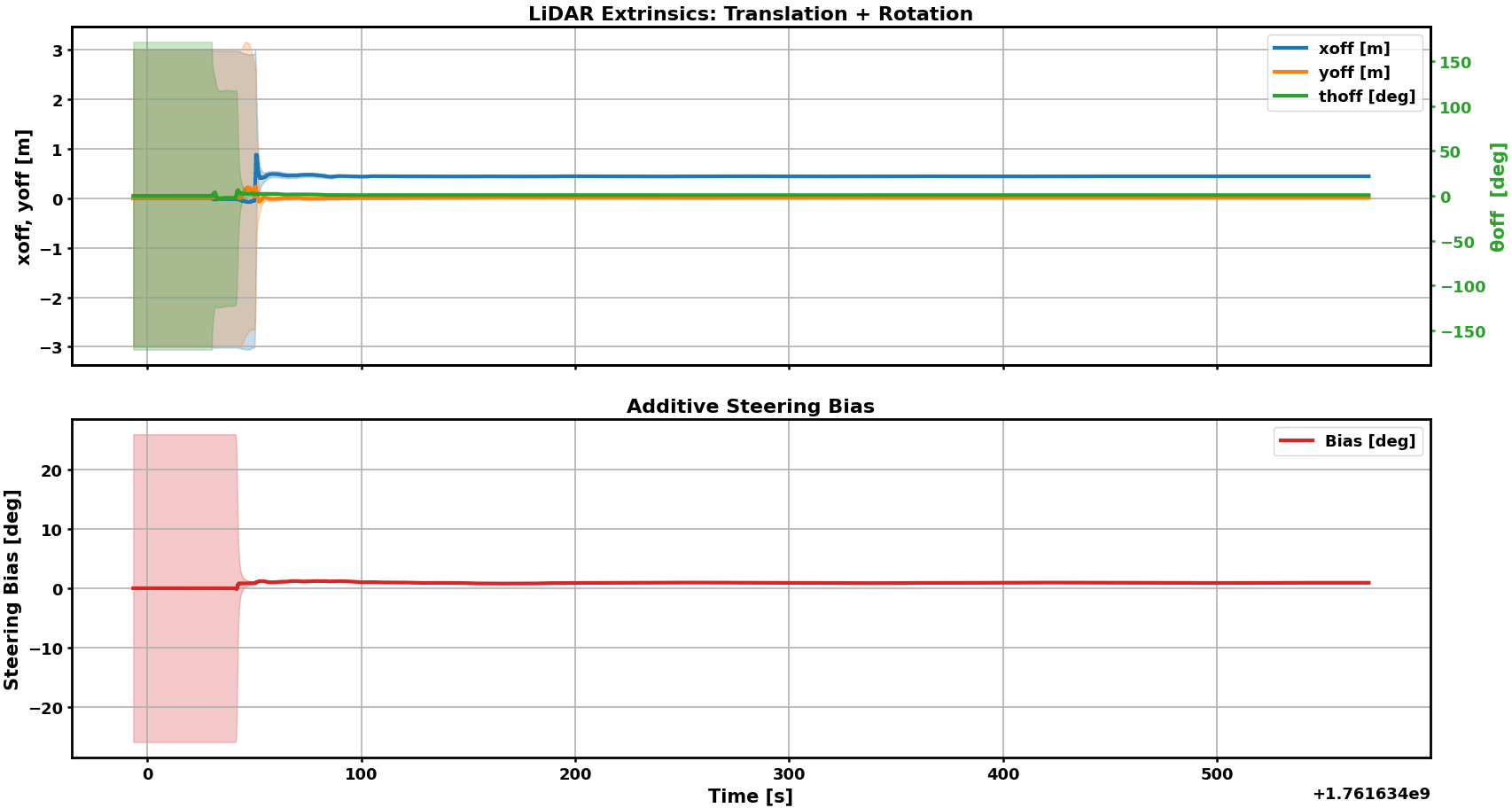}
    }
    \caption{\textbf{Estimator convergence.} With zero initialization, the estimator begins at
    $\mathbf{0}$ and converges to the steady--state values listed in Table~\ref{tab:ladenunladentest_zeroinit}. 
    The initial covariance (faded $3\sigma$ bounds) is intentionally relaxed to allow recovery of the true extrinsic parameters. 
    Note that the $\theta$ axis is on the right.}
    \label{fig:calibxythetadelta_combined_zeroinit}
\vspace{-6mm}
\end{figure}

We first evaluate the estimator on repeated autonomous ``figure-eight'' trajectories (eg. Figure \ref{fig:autonomousrun}) executed under two payload tugging conditions (no load and 500 Kg load). Across both cases, the EKF converges to stable estimates of
$(x_{\mathrm{off}}, y_{\mathrm{off}}, \theta_{\mathrm{off}},  \delta_{off})$. The results are summarized in Table \ref{tab:ladenunladentest} and Figures \ref{fig:calibxythetadelta_unladen} $\&$ \ref{fig:calibxythetadelta_laden}.
\begin{table}[htbp]
    \centering
    \resizebox{\columnwidth}{!}{%
    \begin{tabular}{c c c c c}
        \hline
        \textbf{Trial} 
        & $x_{\text{off}}$ [m] 
        & $y_{\text{off}}$ [m] 
        & $\theta_{\text{off}}$ [°] 
        & $\delta_{\text{off}}$ [°] \\
        \hline
        Unladen          & 0.43 & 0.0046 & 0.72 & 0.92 \\
        Laden (500 kg)   & 0.44 & 0.0055 & 0.73 & 0.92 \\
        \hline
    \end{tabular}%
    }
    \caption{Converged estimates under different load conditions. 
    The known reference for the extrinsics is $[0.434 m,\,0.0,\,0.0]$. Comparison of calibration results for (a) unladen and (b) laden (500\,kg) conditions. 
A consistent $\sim 1\,\text{cm}$ shift in the $x$-component of the estimated extrinsics is observed, 
which may be attributed to LiDAR--odometry noise or load-induced mechanical drift between 
the vehicle body (kinematics frame) and the chassis where the LiDAR is mounted.}
    \label{tab:ladenunladentest}
\vspace{-5mm}
\end{table}
To evaluate the robustness of the estimator in the absence of prior information, we repeat the above experiments with both the LiDAR extrinsics and steering offset initialized to zero, with correspondingly inflated initial covariances. Even in the absence of prior information, the filter reliably converges to the same neighbourhood of steady-state values as observed with tape/CAD-derived initialization, demonstrating that the method remains effective without any prior extrinsic knowledge. The results are reported in Table~\ref{tab:ladenunladentest_zeroinit} and Figure \ref{fig:calibxythetadelta_combined_zeroinit}. A comparison between Table~\ref{tab:ladenunladentest} and Table~\ref{tab:ladenunladentest_zeroinit} shows that both initialization strategies yield closely matching estimates.

Having studied convergence in depth, we do not present convergence plots going forward and only provide the final converged value as results.
\begin{table}[htbp]
    \centering
    \resizebox{\columnwidth}{!}{%
    \begin{tabular}{c c c c c}
        \hline
        \textbf{Trial} 
        & $x_{\text{off}}$ [m] 
        & $y_{\text{off}}$ [m] 
        & $\theta_{\text{off}}$ [°] 
        & $\delta_{\text{off}}$ [°] \\
        \hline
        Unladen          & 0.43 & 0.0052 & 0.72 & 0.91 \\
        Laden (500 kg)   & 0.44 & 0.0056 & 0.73 & 0.91 \\
        \hline
    \end{tabular}%
    }
    \caption{Converged estimates under circumstances similar to \ref{tab:ladenunladentest}. 
    The known reference for the extrinsics is $[0.434 m,\,0.0,\,0.0]$. The filter was intialized to $0$ for all the states under consideration.}
    \label{tab:ladenunladentest_zeroinit}
\vspace{-6mm}
\end{table}
\subsection{Cross-Platform Generalization}
\label{sec: crossplatformgeneralization}
We additionally evaluated the algorithm on a smaller robot from the same platform family (Figure~\ref{fig:atinano}). The estimated LiDAR extrinsics converged to $[0.2597~\mathrm{m},\,0.0101~\mathrm{m},\,-1.29^{\circ}]$ against the reference $[0.26~\mathrm{m},\,0.0~\mathrm{m},\,0.0^{\circ}]$, and the steering bias converged to $-1.11^{\circ}$. These degree-level deviations are consistent with the body-on-frame construction of the platform, where small geometric shifts between the chassis and sensor mount commonly introduce such offsets.
\subsection{Physically induced Steering Offset}
\label{sec: physicallyinducedoffset}
To evaluate the algorithm inte absence of absolute ground truth, we manually introduced steering offsets by physically misaligning the steering mechanism in both positive and negative directions and drove the robot while running the EKF. The filter reliably detected these induced offsets and produced stable estimates of both steering bias and LiDAR extrinsics. As shown in Table~\ref{tab: inducedoffsettest}, the EKF recovers the induced steering offsets with sub-degree accuracy, the worst-case error is $0.27^\circ$ (encoder resolution $0.25^\circ$), and three independent $5^\circ$ trials yield a standard deviation of only $0.03^\circ$. The LiDAR extrinsics remain close to the tape-measured/CAD reference, with $x_{\text{off}}$ varying by $1.7$\,cm, $y_{\text{off}}$ within $3.3$\,cm, and $\theta_{\text{off}}$ within $0.22^\circ$, consistent with scan-matching noise and day-to-day variability due to routine mechanical and electrical testing.
\begin{table}[htbp]
    \centering
    \resizebox{\columnwidth}{!}{%
    \begin{tabular}{c c c c c }
        \hline
        \textbf{Dataset} & Estimated $\delta_{\text{off}}$ [°] 
        & $x_{\text{off}}$ [m] 
        & $y_{\text{off}}$ [m] 
        & $\theta_{\text{off}}$ [°] \\
        \hline
         Induced $\delta_{\text{off}}=-9^\circ$  & $-8.73^\circ$ & 0.4259 & -0.0239 & 0.67 \\
         Induced $\delta_{\text{off}}=7.4^\circ$ & $7.23^\circ$  & 0.4359 & -0.0146 & 0.79 \\
         Induced $\delta_{\text{off}}=5^\circ$   & $4.87^\circ$  & 0.4185 & -0.0034 & 0.88 \\
         Induced $\delta_{\text{off}}=5^\circ$   & $4.87^\circ$  & 0.4212 & 0.0100 & 0.81 \\
         Induced $\delta_{\text{off}}=5^\circ$   & $4.92^\circ$  & 0.4235 & 0.0135 & 0.66 \\
        \hline
    \end{tabular}%
    }
    \caption{Detecting physically induced steering offset. The known reference for the extrinsics is $[0.434 m,\,0.0,\,0.0]$.}
    \label{tab: inducedoffsettest}
\vspace{-5mm}
\end{table}
\subsection{Impact of Steering Offset on Cross Track Error}
\label{sec: impactoncte}
\begin{figure*}[t]
    \centering

    \subfloat[\textcolor{teal}{Runset 1}: Calibration Run\label{fig:a1}]{
        \begin{minipage}[c]{0.32\textwidth}
            \centering
            \includegraphics[width=\linewidth]{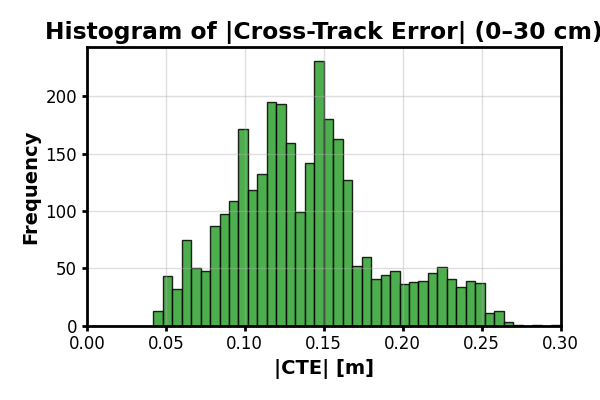}
        \end{minipage}
    }
    \subfloat[\textcolor{brown}{Runset 2}: Calibration Run\label{fig:b1}]{
        \begin{minipage}[c]{0.32\textwidth}
            \centering
            \includegraphics[width=\linewidth]{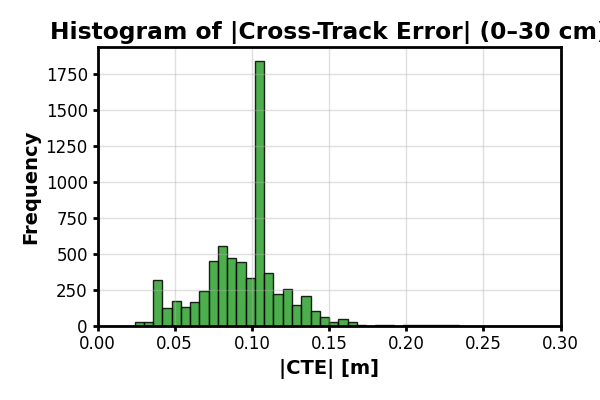}
        \end{minipage}
    }
    \subfloat[\textcolor{violet}{Runset 3}: Calibration Run\label{fig:c1}]{
        \begin{minipage}[c]{0.32\textwidth}
            \centering
            \includegraphics[width=\linewidth]{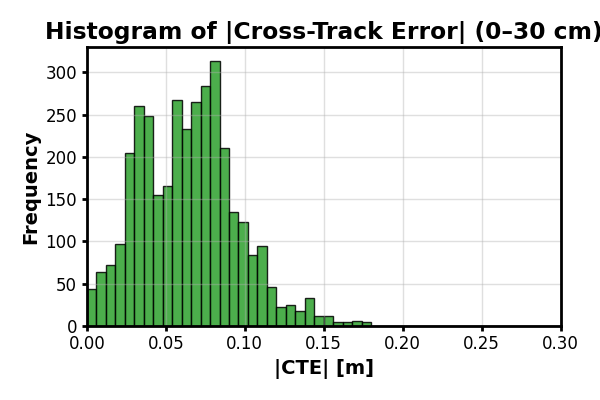}
        \end{minipage}
    }

    \vspace{10pt}

    \subfloat[\textcolor{teal}{Runset 1}: Offset Added\label{fig:a2}]{
        \begin{minipage}[c]{0.32\textwidth}
            \centering
            \includegraphics[width=\linewidth]{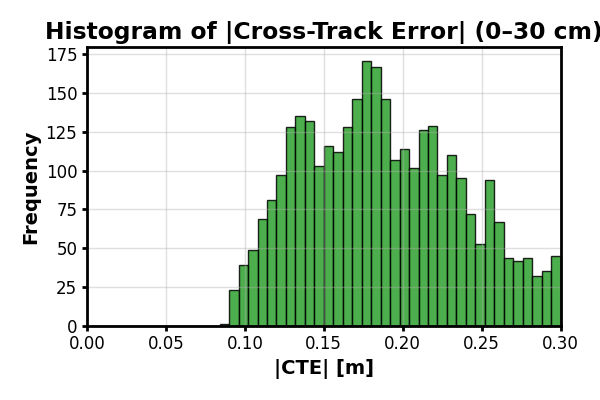}
        \end{minipage}
    }
    \subfloat[\textcolor{brown}{Runset 2}: Offset Added\label{fig:b2}]{
        \begin{minipage}[c]{0.32\textwidth}
            \centering
            \includegraphics[width=\linewidth]{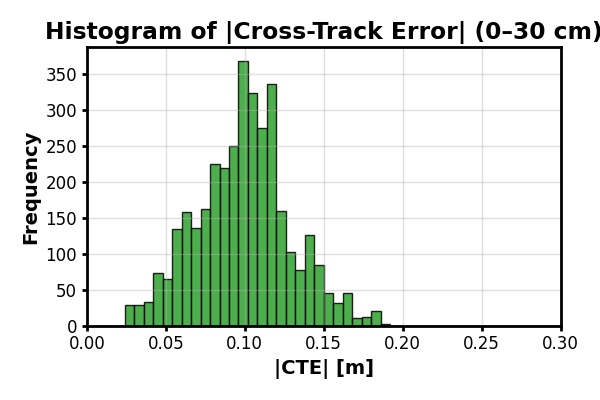}
        \end{minipage}
    }
    \subfloat[\textcolor{violet}{Runset 3}: Offset Added\label{fig:c2}]{
        \begin{minipage}[c]{0.32\textwidth}
            \centering
            \includegraphics[width=\linewidth]{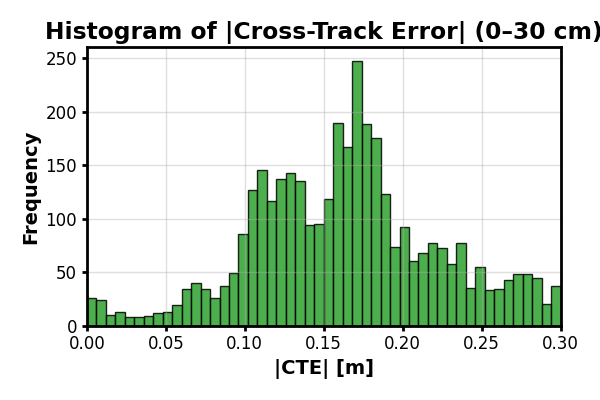}
        \end{minipage}
    }

    \vspace{10pt}

    \subfloat[\textcolor{teal}{Runset 1}: Offset Removed\label{fig:a3}]{
        \begin{minipage}[c]{0.32\textwidth}
            \centering
            \includegraphics[width=\linewidth]{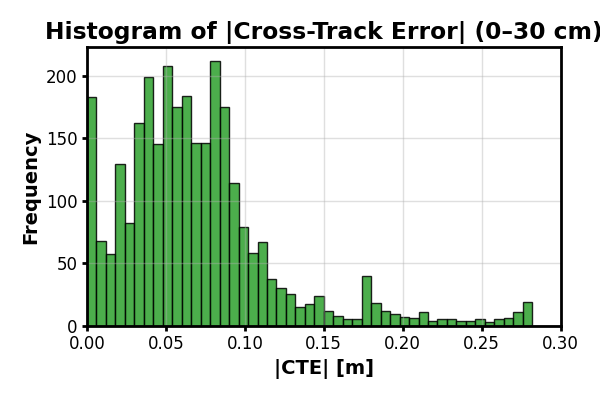}
        \end{minipage}
    }
    \subfloat[\textcolor{brown}{Runset 2}: Offset Removed\label{fig:b3}]{
        \begin{minipage}[c]{0.32\textwidth}
            \centering
            \includegraphics[width=\linewidth]{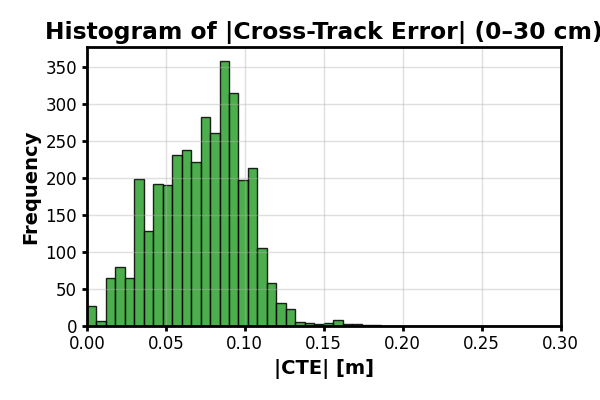}
        \end{minipage}
    }
    \subfloat[\textcolor{violet}{Runset 3}: Offset Removed\label{fig:c3}]{
        \begin{minipage}[c]{0.32\textwidth}
            \centering
            \includegraphics[width=\linewidth]{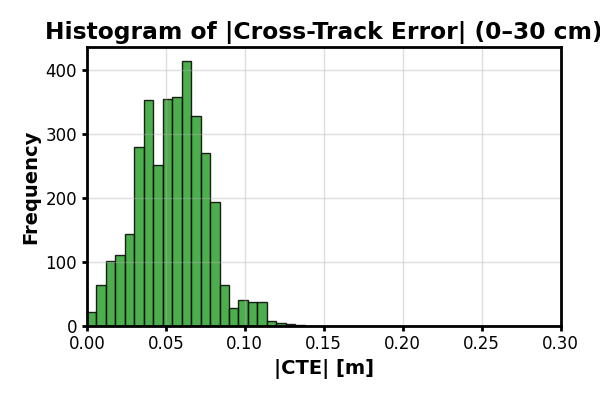}
        \end{minipage}
    }

    \caption{
        Comprehensive evaluation across three experimental conditions:
        (i) calibration run, 
        (ii) physically added steering offset, 
        (iii) physically removed steering offset.
        Columns correspond to Runsets 1–3; rows correspond to the three test scenarios.
    }
    \label{fig:3x3_runs}
\end{figure*}

We evaluated the influence of steering offset on the global CTE, which quantifies how closely the vehicle tracks the planned path. For each trial, we first performed a calibration run to estimate the steering offset, followed by two validation runs: one in which the estimated offset was added to the steering command to exaggerate the miscalibration, and another in which the offset was subtracted to compensate for it. We conducted three such sets of experiments, and the resulting CTE behavior is summarized in Table~\ref{tab:combined_three_days_noday} and visualized in Figure~\ref{fig:3x3_runs}. 

Across all three run sets, adding the estimated steering offset (thus exaggerating the miscalibration) consistently increased the CTE-RMSE, while subtracting it reduced the CTE-RMSE below the calibration baseline (ref. Table \ref{tab:combined_three_days_noday}). The worst-case degradation occurred in Run Set~3, where CTE-RMSE rose from $0.071$\,m to $0.174$\,m, whereas compensation lowered it to $0.059$\,m. The smaller-offset cases (Run Sets~1 and~2) exhibited proportionally smaller improvements, reflecting the direct relationship between the degree of miscalibration and the achievable reduction in tracking error. These trials were conducted over several days of normal robot operation, yet the LiDAR extrinsics remained stable (ranges: $x_{\text{off}}$ within $0.5$\,cm, $y_{\text{off}}$ within $1$\,cm, and $\theta_{\text{off}}$ within $0.08^\circ$). Overall, the results demonstrate that steering offset directly impacts the CTE and eliminating the offset improves path-tracking performance.

While the CTE–RMSE values in Table~\ref{tab:combined_three_days_noday} summarize the overall trend, they compress a system-level error influenced by controller tuning, wheel slip, floor conditions, and transient disturbances. Thus, modest 10--20\% changes in RMSE may not fully expose the effect of miscalibration. Moreover, these tests were performed in a constrained indoor area allowing only a few tens of meters of motion; in real deployments with trajectories spanning hundreds of meters, the impact of steering bias becomes far more pronounced.
\begin{table}[htbp]
    \centering
    \resizebox{\columnwidth}{!}{%
    \begin{tabular}{c c c c c c}
        \hline
        \textbf{Run} &
        $\delta_{\text{off}}~[^\circ]$ &
        $x_{\text{off}}$ [m] &
        $y_{\text{off}}$ [m] &
        $\theta_{\text{off}}~[^\circ]$ &
        RMSE-CTE [m] \\
        \hline

        \multicolumn{6}{l}{\textcolor{teal}{\textbf{Run Set 1}}} \\
        Calibration run 
        & -1.25 & 0.4393 & -0.0088 & 0.71 & \textcolor{orange}{0.14} \\
        $\delta_{\text{off}} = -1.25^\circ$ added 
        & -2.53 & 0.4383 & -0.0132 & 0.68 & \textcolor{red}{0.20} \\
        $\delta_{\text{off}} = -1.25^\circ$ subtracted
        & -0.07 & 0.4394 & -0.0119 & 0.69 & \textcolor{blue}{0.09} \\
        \midrule

        \multicolumn{6}{l}{\textcolor{brown}{\textbf{Run Set 2}}} \\
        Calibration run 
        & -0.41 & 0.4356 & -0.0158 & 0.75 & \textcolor{orange}{0.09} \\
        $\delta_{\text{off}} = -0.41^\circ$ added
        & -0.78 & 0.4400 & -0.0184 & 0.71 & \textcolor{red}{0.10} \\
        $\delta_{\text{off}} = -0.41^\circ$ subtracted
        & -0.01 & 0.4400 & -0.0190 & 0.75 & \textcolor{blue}{0.07} \\
        \midrule

        \multicolumn{6}{l}{\textcolor{violet}{\textbf{Run Set 3}}} \\
        Calibration run 
        & 2.92 & 0.4382 & -0.0158 & 0.76 & \textcolor{orange}{0.071} \\
        $\delta_{\text{off}} = 2.92^\circ$ added
        & 5.84 & 0.4357 & -0.0140 & 0.76 & \textcolor{red}{0.174} \\
        $\delta_{\text{off}} = 2.92^\circ$ subtracted
        & -0.03 & 0.4382 & -0.0158 & 0.76 & \textcolor{blue}{0.059} \\
        \hline
    \end{tabular}
    }
    \caption{Effect of Estimated Steering Offset on Cross-Track Error (CTE) Across Three Experimental Run Sets. The known reference for the extrinsics is $[0.434 m,\,0.0,\,0.0]$. The topmost row contains labels for the estimated steering offset, the extrinsics, and RMSE-CTE.}
    \label{tab:combined_three_days_noday}
    \vspace{-7mm}
\end{table}

To provide a more complete picture, Figure \ref{fig:3x3_runs} visualizes the full distribution of absolute CTE for each run. These histograms show consistent shifts: adding the estimated steering offset yields broader, right-shifted, heavy-tailed distributions indicative of exaggerated curvature errors, whereas subtracting the offset produces concentrated distributions with few large-error outliers. This distribution-level view confirms that the estimated steering offset induces a systematic and physically meaningful change in path-tracking behavior that cannot be fully captured by the RMSE alone.
\begin{figure*}[htbp]
    \centering

    \subfloat[Manual teleoperation run showing EKF trajectory, 
              uncompensated bicycle propagation, and bias-corrected propagation. 
              The RMSE between EKF and uncompensated trajectory is 
              \textcolor{red}{15.316} m, whereas between EKF and compensated 
              trajectory is \textcolor{blue}{1.359} m.\label{fig:manualrun}]{
        \begin{minipage}[t]{0.48\textwidth}
            \centering
            \includegraphics[width=0.8\linewidth]{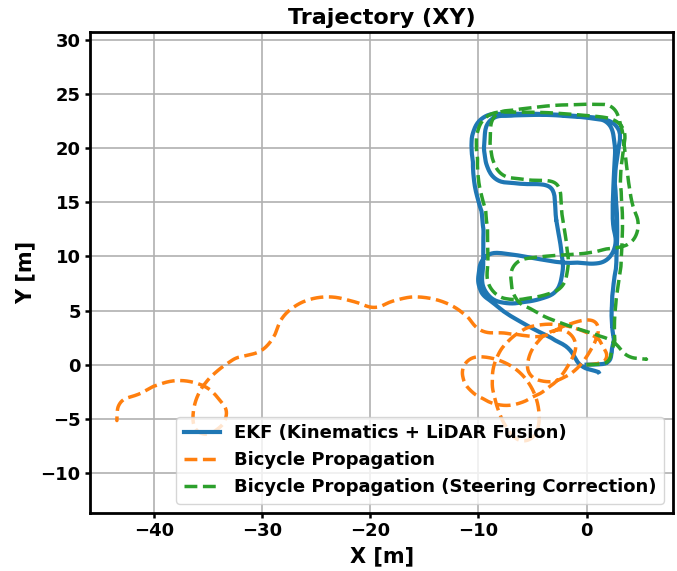}
        \end{minipage}
    }
    \hfill
    \subfloat[Autonomous run showing a similar comparison between EKF, 
              uncompensated, and bias-corrected propagation. 
              The RMSE between EKF and uncompensated trajectory is 
              \textcolor{red}{14.02} m, whereas between EKF and compensated 
              trajectory is \textcolor{blue}{1.65} m.\label{fig:autonomousrun}]{
        \begin{minipage}[t]{0.48\textwidth}
            \centering
            \includegraphics[width=0.8\linewidth]{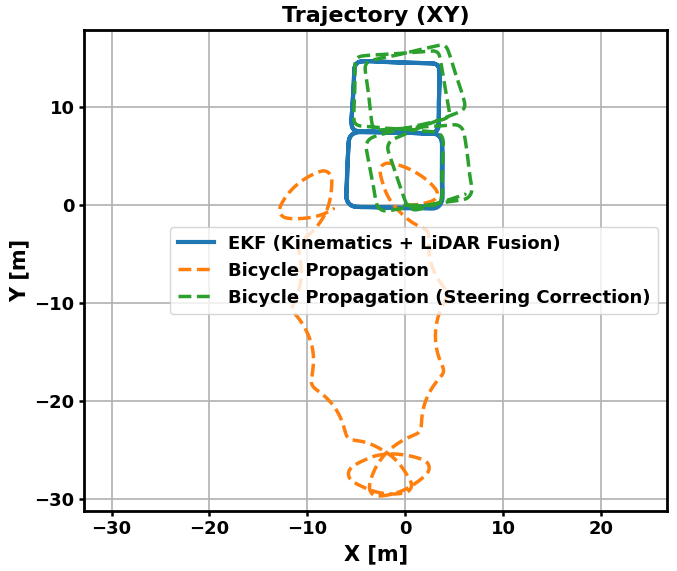}
        \end{minipage}
    }

    \caption{Comparison of EKF-estimated trajectory with uncompensated and 
    steering-bias-corrected bicycle propagation across 
    (a) manual teleoperation 
    and (b) autonomous navigation trials. 
    In both cases, the compensated model closely follows the EKF estimate, 
    whereas the uncompensated model exhibits significant drift.}
    \label{fig:manualvsautonomousrun}
    \vspace{-8mm}
\end{figure*}

\subsection{Impact of Steering Offset on Odometry}
\label{sec: impactonodometry}

Figure~\ref{fig:manualvsautonomousrun} compares the trajectory traces for two
experiments: manual teleoperation (Figure~\ref{fig:manualrun}) and autonomous
navigation (Figure~\ref{fig:autonomousrun}). In both cases, the
\emph{uncompensated} bicycle model diverges rapidly from the true motion, whereas
the \emph{bias-compensated} model remains tightly aligned with the EKF estimate.
This indicates that even in the absence of LiDAR-odometry updates such as in
narrow corridors or low-structure regions where LiDAR odometry often fails the
compensated bicycle propagation provides a reliable standalone local state estimate.
\vspace{-4mm}
\section{Conclusion and Future Work}
In this work, we presented an online automatic method to jointly estimate the planar
LiDAR extrinsics and the steering-angle offset of a bicycle-type wheeled mobile
robot using an Extended Kalman Filter whose measurement model uses LiDAR-odometry
pose under the motion-based calibration framework~\cite{taylor2016} to jointly
estimate vehicle pose and calibration parameters. Although LiDAR odometry is used
as the pose source in this implementation, the modular EKF structure \cite{lynen2013} allows any
pose-generating sensor (e.g., visual odometry, GPS) to be
plugged in, enabling steering and extrinsic calibration across different sensing stacks. The
proposed formulation enables continual estimation of calibration, thereby
improving the accuracy of the propagated vehicle state (Figure~\ref{fig:manualvsautonomousrun}). We demonstrated the convergence properties of the estimator across
multiple experimental conditions, including evaluations on two distinct platforms
with different mechanical dimensions. In the absence of ground truth, we
physically induced steering offsets to validate that the filter can reliably track
changes in real time. Furthermore, we evaluated the influence of the estimated
steering offset on downstream control metrics, particularly cross-track error
(CTE), and provided a statistical analysis of how steering bias affects the
distribution of absolute CTE. Across all experiments, the estimated LiDAR
extrinsics consistently converged to values close to the expected reference. Future work will investigate four main directions. First, we aim to deploy the
filter fully online on the robot and perform closed-loop evaluations where
steering-offset correction is applied continuously within the controller. Second,
we plan to study the sensitivity of the system to LiDAR extrinsics, including
their impact on localization and control in more complex environments. Third, we
will extend the EKF formulation to jointly refine key vehicle parameters such as
wheelbase, wheel radius, and track width to account for model uncertainty and
mechanical drift. Finally, we will conduct a detailed observability analysis of
the joint calibration problem to identify informative excitation conditions and
design trajectories that ensure all states remain well observable.

\bibliographystyle{IEEEtran}
\vspace{-4mm}
\bibliography{refs}
\end{document}